\pdfoutput=1

\documentclass[11pt]{article}

\usepackage[preprint]{acl}
\usepackage{listings}
\usepackage{times}
\usepackage{latexsym}

\usepackage{booktabs}    
\usepackage{multirow}    
\usepackage{amsmath}
\usepackage{float}
\usepackage{colortbl}  
\usepackage{amssymb}

\usepackage[T1]{fontenc}

\usepackage[utf8]{inputenc}

\usepackage{microtype}

\usepackage{inconsolata}

\usepackage{graphicx}

\definecolor{Gray}{gray}{0.9} 
\newcommand{\percentred}[1]{\textcolor{red}{\scriptsize #1}}
\newcommand{\percentgreen}[1]{\textcolor{green}{\scriptsize #1}}

\title{Mitigating Hallucinations in Large Vision-Language Models via Entity-Centric Multimodal Preference Optimization}

\author{
  Jiulong Wu$^{1}$,
  Zhengliang Shi$^{3}$, 
  Shuaiqiang Wang$^{2}$,
  Jizhou Huang$^{2}$ \\
  {\bf Dawei Yin}$^{2}$,
  {\bf Lingyong Yan}$^{2}$$^*$,
  {\bf Min Cao}$^{1}$$^*$,
  {\bf Min Zhang}$^{1}$ \\
  $^{1}$Soochow University, Suzhou, China \ \ 
  $^{2}$Baidu Inc., Beijing, China \\
  $^{3}$Shandong University, Qingdao, China \\
  \texttt{wjlwujiulong@gmail.com, zhengliang.shii@gmail.com},\\ 
  \texttt{lingyongy@gmail.com, mcao@suda.edu.cn}
}

\begin{document}
\maketitle

\def\thefootnote{*}\footnotetext{Co-corresponding authors.}

\begin{abstract}
Large Visual Language Models (LVLMs) have demonstrated impressive capabilities across multiple tasks.
However, their trustworthiness is often challenged by hallucinations, which can be attributed to the modality misalignment and the inherent hallucinations of their underlying Large Language Models (LLMs) backbone.
Existing preference alignment methods focus on aligning model responses with human preferences while neglecting image-text modality alignment, resulting in over-reliance on LLMs and hallucinations.
In this paper, we propose Entity-centric Multimodal Preference Optimization (EMPO), which achieves enhanced modality alignment compared to existing human preference alignment methods.
Besides, to overcome the scarcity of high-quality multimodal preference data, we utilize open-source instruction datasets to automatically construct high-quality preference data across three aspects: image, instruction, and response.
Experiments on two human preference datasets and five multimodal hallucination benchmarks demonstrate the effectiveness of EMPO, e.g., reducing hallucination rates by 85.9\% on Object-HalBench and 49.8\% on MM-HalBench.
The code and dataset will be released at \url{https://github.com/RobitsG/EMPO}.
\end{abstract}    
\section{Introduction}
\label{sec:intro}

Large Vision Language Models (LVLMs) have recently demonstrated impressive capabilities in multimodal question answering~\cite{chen2023minigpt,liu2023llava,qwenvl,lu2024deepseek,lyu2025deepshop}, which typically consists of a visual encoder to extract image features, and a Large Language Model (LLM) to answer the image-related textual instructions based on the provided visual context.
The LVLMs are usually learned in two steps~\cite{li2023blip,du2022survey,lin2024vila}: (1) pretraining on large-scale image-text pairs to learn multimodal knowledge, and (2) fine-tuning on multimodal instruction-following datasets to steer their responsiveness to user instructions~\cite{liu2023llava,wang2024visionllm, qwenvl}.

\begin{figure}
    \setlength{\abovecaptionskip}{4pt}
    \centering
    \includegraphics[width=1\linewidth]{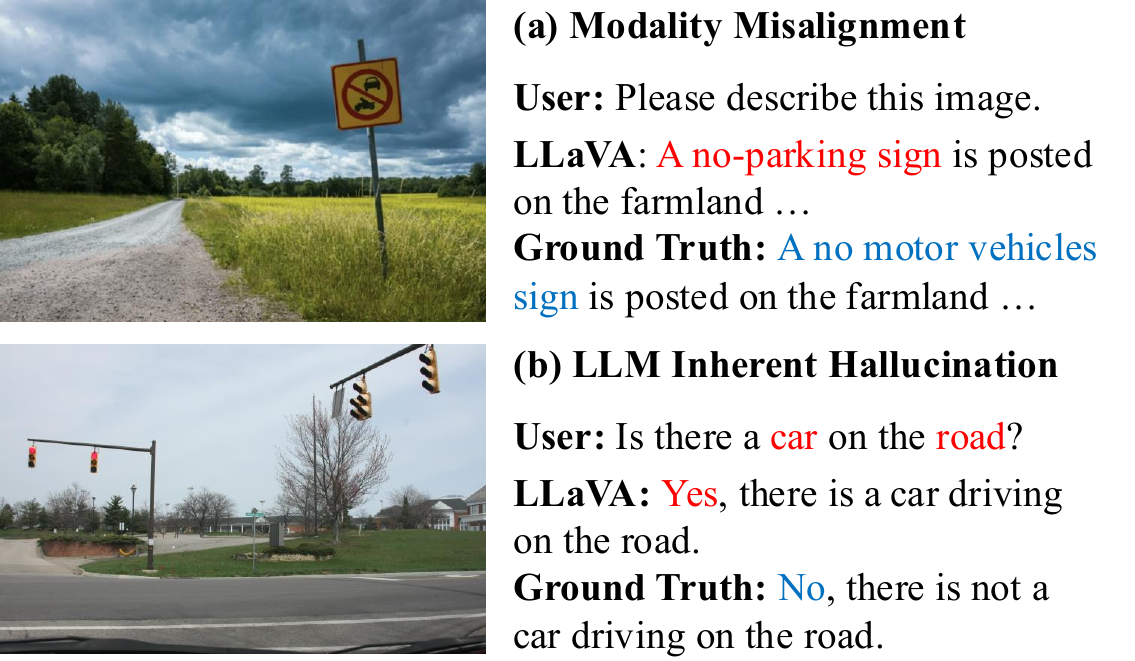}
    \caption{Two types of hallucinations.  
    a) Modality misalignment: LVLM recognizes the presence of entities but confuses their semantics (e.g., misinterpreting a sign as "No Parking" instead of "No Motor Vehicles Allowed").
    b) LLM inherent hallucination: The LVLM's response is entirely dependent on textual context, disregarding the image content  (e.g., inferring a "car" must be present whenever a "road" is present).}
    \label{fig:cause}
    \vspace{-12pt}
\end{figure}

However, recent studies have identified that the LVLMs usually suffer from the hallucination problem~\cite{li2023evaluating,liu2024survey,gunjal2024detecting,guan2024hallusionbench,jiang2024hallucination,jiang2024hal}, akin to the LLM hallucinations~\cite{zhang2023siren,li2023halueval,dhuliawala2023chain}. 
Specifically, there are usually two types of LVLM hallucinations~(as shown in Figure~\ref{fig:cause})~\cite{liu2024survey,lan2024survey}.
The first type is the \textbf{modality misalignment}, which arises from the modality gap between the visual encoder and LLM, resulting in semantic mismatches between image context and textual instructions. 
For instance, in Figure~\ref{fig:cause}(a), the LVLM (i.e., LLaVa)~\cite{liu2023llava} correctly identifies the sign on the farmland but misinterprets its meaning as \emph{"No Parking"} instead of the correct interpretation, \emph{"No Motor Vehicles Allowed"}.
The second type is the \textbf{LLM inherent hallucination}.
When the LLM inherent knowledge is either incorrect or conflicts with visual inputs, hallucinations manifest as entity co-occurrence phenomena~\cite{lan2024survey}. 
For example, as shown in Figure~\ref{fig:cause}(b), \emph{"car"} and \emph{"road"} frequently co-occur in the LLM's pretraining corpus; the LVLM erroneously infers that whenever a \emph{"road"} is present, a \emph{"car"} must also be present, disregarding the image content.

To mitigate hallucinations in LVLMs, many recent studies~\cite{povid, llavarlhf, yu2024rlaif} adopt preference alignment algorithms such as Direct Preference Optimization (DPO)~\cite{rafailov2024direct}, to align the model's multimodal response capabilities with human preferences. However, existing multimodal preference optimization methods extend DPO by simply adding images to preference conditions, without paying sufficient attention to entity-centric factual fragments, which are highly related to hallucinations.
On the other hand, incorporating new modalities into preference conditions increases the range of possible preference combinations, necessitating a comprehensive exploration of preference dimensions to ensure LVLM responses effectively overcome the inherent hallucinations of LLMs based on images and user instructions.

To address these questions, we propose Entity-Centric Multimodal Preference Optimization (EMPO), which efficiently aligns images, user instructions, and model responses through preference optimization on comprehensive aspects. 
Specifically, we first construct a multimodal preference dataset based on open-source image-text instruction datasets~\cite{povid, yu2024rlaif} by automatically editing the entities, attributes, and relationships within the image-instruction-response triplets. 
Then, we apply the DPO loss across three aspects—image instruction conditioning and model responses, helping LVLMs align image entity features with the corresponding textual semantics in user instructions and model responses.
To validate the effectiveness of EMPO, we evaluate it on five benchmarks under the LLaVA-1.5~\cite{li2024llava} framework with two training datasets. Experimental results demonstrate that EMPO achieves lower hallucination rates than GPT-4V~\cite{chen2024sharegpt4v} across three hallucination benchmarks. Furthermore, compared to the well-known DPO algorithm, EMPO reduces hallucination rates by 85.9\% on Object HalBench~\cite{rohrbach2018chair} and by 49.8\% on MM HalBench~\cite{llavarlhf}.

We summarize our contributions as:
(1) We propose EMPO to effectively mitigate hallucinations by addressing the insufficient alignment of image and text modalities in existing multimodal DPO algorithms. Through preference optimization across three aspects, EMPO helps LVLMs align entity features and semantic concepts better.
(2) We propose automatically constructing multimodal preference data using open-source instruction datasets to address the lack of high-quality multimodal preference data caused by the increasing complexity of preference combinations. Our data construction method can be applied to any existing instruction dataset without additional manual annotation.
(3) Experimental results on two preference training datasets across five widely-used benchmarks show that EMPO enhances multimodal semantic alignment and effectively reduces hallucinations.
\section{Related Work}
\label{sec:related_work}

\paragraph{Large Vision Language Models.}
Recent research on LVLMs~\cite{zhu2023minigpt,liu2023llava,qwenvl,lu2024deepseek,zhang2024internlm,xu2025efficienttext2video} constructs LVLMs by aligning LLMs with visual models, demonstrating superior performance across various visual–language tasks compared to earlier studies~\cite{jia2021scaling,radford2021learning,ju2022prompting,alayrac2022flamingo}. 
These LVLMs typically adopt a two-stage training strategy: 
(1) \textbf{Pretraining} on large-scale image–text pairs to learn fundamental multimodal knowledge~\cite{li2023blip,du2022survey,lin2024vila,qwenvl}, and 
(2) \textbf{Instruction fine-tuning} using instruction datasets to improve instruction-following abilities~\cite{chen2024your,wang2024visionllm,qwenvl,wang2024vigc,li2023blip,li2024llava}. 
For instance, LLaVA~\cite{li2024llava} introduces synthetic instructions to fine-tune an instruction-following LVLM, while MiniGPT-v2~\cite{chen2023minigpt} employs unique task identifiers during fine-tuning to reduce instruction ambiguity.

\paragraph{Hallucination in LVLMs.}
Despite their impressive performance, LVLMs suffer from \emph{hallucinations}, where model responses conflict with the images, instructions, or context~\cite{du2022survey,llavarlhf,xiao2025alibabaai}. 
To mitigate the hallucination, some methods have been proposed to filter out long-tail or entity co-occurrence data~\cite{liu2023mllms,yu2024hallucidoctor,hu2023ciem,liu2023mitigating}, though this involves high annotation costs. 
Others recognized modal misalignment as a key factor~\cite{li2023factual,tong2024eyes,cao2024dualfocus,jiang2024joint,jiang2024hallucination}, yet overlooked inherent LLM errors.
Post-processing techniques—optimizing decoding strategies~\cite{zhang2025self,huang2024opera,yang2024pensieve,gao2024fact,Leng_2024_CVPR_VCD} or applying post-hoc corrections~\cite{lee2023volcano,zhou2023analyzing,yin2023woodpecker}—reduce hallucinations but add inference cost. 
Human preference alignment has also emerged as an effective approach to mitigate hallucinations~\cite{lan2024survey}. 
LLaVA-RLHF~\cite{llavarlhf} pioneered this exploration in LVLMs, while RLHF-V~\cite{yu2024rlhf}, RLAIF-V~\cite{yu2024rlaif}, and POVID~\cite{povid} further refined the approach with improved visual localization, text segment scoring, and preference example generation. 
However, these methods primarily focus on response-level preferences while neglecting the multimodal task's requirements to align human preferences with images and instructions in multiple aspects. MDPO~\cite{wang2024mdpo} proposed image-conditional preference alignment but overlooked aligning instructions with human preferences.
In contrast, our EMPO incorporates preferences across comprehensive aspects
for efficient alignment between visual content and semantic concepts.
\begin{figure*}
    \setlength{\abovecaptionskip}{0pt}
    \setlength{\belowcaptionskip}{-10pt}
    \centering    
    \includegraphics[width=1\linewidth]{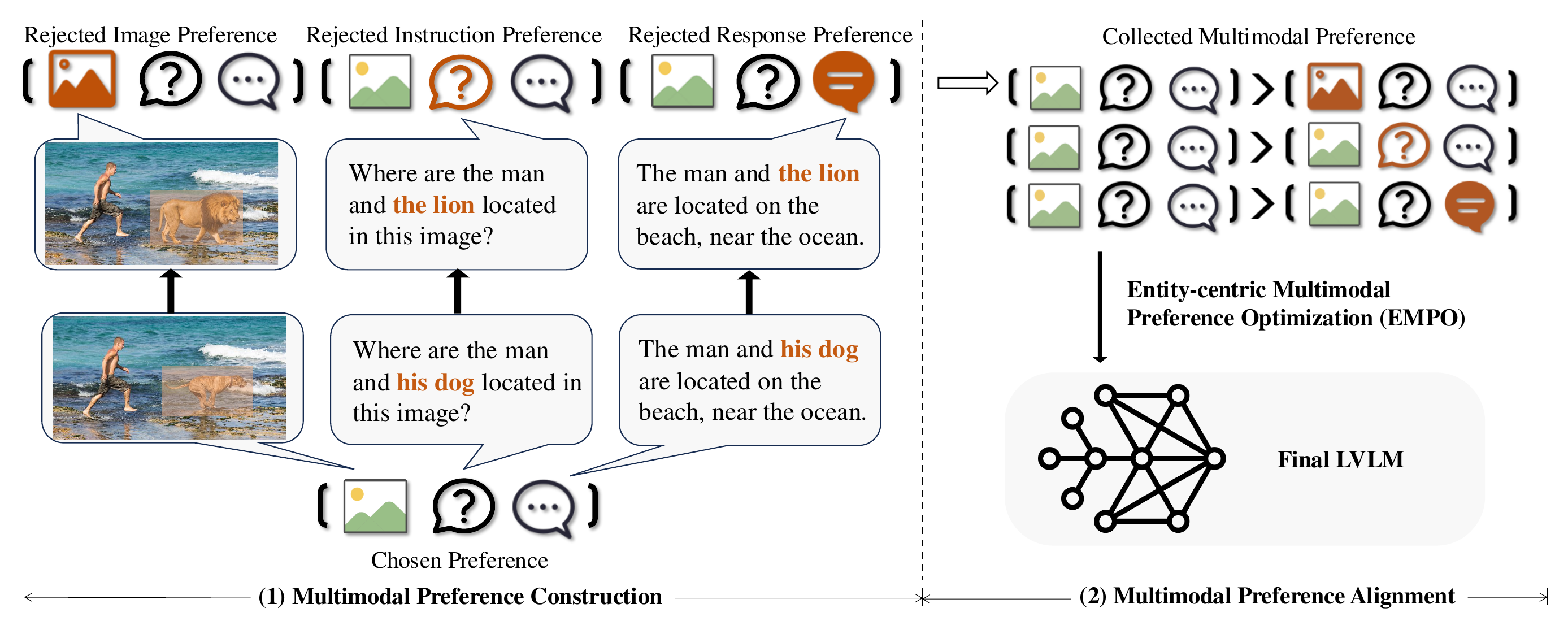}
    \caption{Illustration of our framework. (1) At the data level, we construct a fine-grained preference alignment dataset across three aspects: image, instruction condition, and model response. (2) At the method level, we propose entity-centric multimodal preference optimization for aligning image contents with semantic concepts.}
    \label{fig:framework}
\end{figure*}

\section{Method}
\label{sec:method}
In this section, we detail the data construction and training process in the proposed EMPO.
Section~\ref{sec:preliminary} introduces the foundation of preference optimization.
Section~\ref{sec:dataset} details the procedure to construct multimodal preference data.
Section~\ref{sec:loss} introduces how EMPO addresses the hallucination by an entity-centric alignment training framework.

\subsection{Preliminaries}
\label{sec:preliminary}
Direct Preference Optimization (DPO)~\cite{rafailov2024direct} is a primary method for human preference alignment that implicitly models the reward function in Reinforcement Learning from Human Feedback~\cite{yu2024rlhf}. 
By training on preference pairs that distinguish between accurate and hallucinated responses, DPO reduces the model's tendency to produce hallucinations.
Given an instruction $q$, chosen response $y_w$, and rejected response $y_l$, DPO formulates the reward function as:
\begin{equation}
    r(q, y)=\beta \log \frac{\pi_{\theta}(y \mid q)}{\pi_{\mathrm{ref}}(y \mid q)}+Z(q),
\end{equation}   
where $Z(q)$ is a partition function, $\pi_{\theta}$ represents the policy model, $\pi_{ref}$ is the reference model, and $\beta$ is a hyperparameter controlling deviation from the reference model. DPO directly optimizes:
\begin{equation}
\small
\begin{aligned}
    \mathcal{L}_{\mathrm{DPO}} = -\log\sigma( r(q, y_w) &- r(q, y_l) )\\
    = -\log\sigma( \beta \log \frac{\pi_{\theta}\left(y_{w} \mid q\right)}{\pi_{\mathrm{ref}}\left(y_{w} \mid q\right)} &- \beta \log \frac{\pi_{\theta}\left(y_{l} \mid q\right)}{\pi_{\mathrm{ref}}\left(y_{l} \mid q\right)} ).
\end{aligned}
\end{equation}

\subsection{Preference Dataset Construction}\label{sec:dataset}
As aforementioned, most existing multimodal preference datasets~\cite{povid, yu2024rlaif} merely align human preferences with the overall response, lacking focus on entity-centered key facts in the images and instructions.
Therefore, we perform comprehensive multimodal alignment via constructing multimodal preference datasets across the full aspects: image context, instruction condition, and model response.
As shown in Figure~\ref{fig:framework}, we keep the original images, instructions, and response data as chosen samples, and then edit the entities, attributes, and relationships in three aspects respectively to construct rejected preference samples. 

\paragraph{Image Preference Data}
\label{sec:image preference}
To construct the rejected image, we initially employ GPT4o-mini~\cite{achiam2023gpt} to identify entities in both the instruction and response, ensuring close alignment between the edited image and text. 
Subsequently, we use an object detection model to locate these entities.  
Next, we apply Stable-diffusion-2~(SD2)~\cite{sd2} to either remove 30\% entities or substitute them with visually plausible alternatives, thereby generating an edited image as a rejected image sample \(v_l\).
Finally, we use CLIP~\cite{jiang2023clip} to calculate the similarity between the edited image regions and the entity labels to ensure the image has been correctly edited.
Based on the rejected image samples, these selected entities will be weighted as described in Section~\ref{sec:loss}.
We introduce two strategies to construct image preference rejected samples \( q_l \):
(1) \textit{Entity Deleting}: Use SD2 to delete the chosen entities, helping the LVLM reduce the occurrence of non-existent entities generated by the LVLM. 
(2) \textit{Entity Replacement}: Use SD2 to replace the chosen entities with incorrect but high-frequency entities, helping the LVLM overcome entity co-occurrence hallucinations~\cite{du2022survey}.

\paragraph{Instruction Preference Data}
\label{sec:instruction preference}
We employ GPT4o-mini~\cite{achiam2023gpt} to adapt the original instructions in positions of the selected entities above, as well as their related attributes and relationships, thereby constructing rejected instructions \(q_l\). 
Consistent with the findings of HA-DPO~\cite{ha-dpo}, we observe that the distribution of GPT-modified instructions differs from the vanilla instructions, resulting in a decline in performance. 
We analyze that rejected samples serve as hard negative samples~\cite{kalantidis2020hard} because they are sufficiently similar to chosen samples while highlighting hallucination-related factual errors, which enhances LVLM's attention to entity-centric key facts.
We use word2vec~\cite{mikolov2013word2vec} and rule-based matching to ensure that the rejected instruction \(q_l\) and chosen instruction \(q_w\) have different semantic meanings but the same syntactic structure.

\paragraph{Response Preference Data}
\label{sec:response preference}
We propose a general multimodal preference data construction method applicable to any instruction dataset, demonstrating it by construct response preferences from two open-source datasets.
The first is POVID~\cite{povid} containing 17,332 entries, where we collect the rejected image preference sample $(v_l)$ and the rejected instruction preference sample $(q_l)$ from the previous paragraphs, using them as LVLM input to generate incorrect responses as the rejected response preference sample $(y_l)$.
The second is RLAIF-V~\cite{yu2024rlaif}, containing 81,342 entries, where we employed MiniCPM-V2.5~\cite{yao2024minicpm} to compare two candidate answers generated by LLaVA-1.5~\cite{li2024llava}, establishing preference rankings between responses in four iterations.
Human evaluation conducted by experts on randomly sampled entries confirms the quality of these datasets. 
Due to space constraints, quality control and cost analysis details are included in Appendix~\ref{app:data_quality}.

\subsection{Entity-centric Multimodal Preference Optimization}
\label{sec:loss}
In the context of LVLMs, aligning human preference includes three aspects: image, instruction, and response. 
To enhance LVLM's attention to image features and mitigate hallucinations caused by over-reliance on text modality, we align image and text modalities through explicit preference optimization of image instruction conditions and model responses. 
We define three optimization objectives: $\mathcal{L}_{\mathrm{DPOv}}$ for improving visual entity recognition, $\mathcal{L}_{\mathrm{DPOq}}$ for enhancing instruction following, and $\mathcal{L}_{\mathrm{DPOr}}$ directly align with human preference:
\begin{equation}
\begin{aligned}
    \mathcal{L}_{\mathrm{DPOv}} = -\log \sigma (&\beta \log \frac{\pi_{\theta}\left(y \mid v_w, q\right)}{\pi_{\mathrm{ref}}\left(y \mid v_w, q\right)} \\ 
    - &\beta \log \frac{\pi_{\theta}\left(y \mid v_l, q\right)}{\pi_{\mathrm{ref}}\left(y \mid v_l, q\right)}),
\end{aligned}  
\label{eq3}
\end{equation}
\begin{equation}
\begin{aligned}
    \mathcal{L}_{\mathrm{DPOq}} = -\log \sigma(&\beta \log \frac{\pi_{\theta}\left(y \mid v, q_w\right)}{\pi_{\mathrm{ref}}\left(y \mid v, q_w\right)} \\
   \ - &\beta \log \frac{\pi_{\theta}\left(y \mid v, q_l\right)}{\pi_{\mathrm{ref}}\left(y \mid v, q_l\right)}),
\end{aligned}
\label{eq4}
\end{equation}
\begin{equation}
\begin{aligned}
    \mathcal{L}_{\mathrm{DPOr}} =-\log \sigma(&\beta \log \frac{\pi_{\theta}\left(y_w \mid v, q\right)} {\pi_{\mathrm{ref}}\left(y_w \mid v, q\right)} \\
    -&\beta \log \frac{\pi_{\theta}\left(y_l \mid v, q\right)}{\pi_{\mathrm{ref}}\left(y_l \mid v, q\right)}),
\end{aligned}
\label{eq5}
\end{equation}
where $w$ and $l$ represent chosen and rejected preferences respectively, and $v = v_w$, $q = q_w$, $y = y_w$.

Introducing image conditions brings about more complex preference combinations, making it challenging to use vanilla DPO to assign credit to desirable key facts, leading to reward hacking~\cite{pan2024hacking}. We propose assigning token weights to key entities in three aspects to solve this problem.
Specifically, we allocate higher weights to critical features in the image, instruction, and response, thereby enhancing the LVLM's focus on entity features and enabling it to distinguish hallucinated tokens from non-hallucinated ones better. 
It is noteworthy that assigning token weights does not incur additional effort, as the positions of high-weight tokens are already determined during the construction of preference data. 
The formula for assigning weights to model outputs is as follows:
\begin{equation}
\small
\begin{aligned}
    \log \pi(y \mid v, q)=(1-\alpha )\sum_{y_{i} \notin y_{e}} \log p\left(y_{i} \mid v, q, y_{<i}\right)\\
    +\alpha \sum_{y_{i} \in y_{e}} \log p\left(y_{i} \mid v, q, y_{<i}\right),
\end{aligned} 
\label{fomula: weight}
\end{equation}    
where $\alpha$ is a weighting hyperparameter, $y_i$ is the $i$-th token of response $y$, with larger $\alpha$ indicating greater token influence on preference. In this way, emphasizing hallucination-related entities strengthens human preference feedback to the LVLM, thereby enhancing its factual accuracy.

The overall multimodal preference optimization objective combines all three aspects:
\begin{equation}
    \mathcal{L}_{\mathrm{EMPO}} = \mathcal{L}_{\mathrm{DPOv}} +
    \mathcal{L}_{\mathrm{DPOq}} + \mathcal{L}_{\mathrm{DPOr}},
\label{fomula: final}
\end{equation}
where the LVLMs are optimized to fully align hallucination-related key facts with human preference, reducing their hallucinations. 
\section{Experiments}
\label{sec:experiments}
\begin{table*}[t] 
    \resizebox{\linewidth}{!}{  
    \begin{tabular}{l c c cc c cc cc}  
    \toprule  

    \multirow{2}{*}{\textbf{Model}}  & \multirow{2}{*}{\textbf{Model}} & \multirow{2}{*}{\textbf{Size}} & \multicolumn{2}{c}{\textbf{Object-HalBench}} & \multicolumn{2}{c}{\textbf{MMHal-Bench}} & \multicolumn{2}{c}{\textbf{AMBER}} \\
    
    \cmidrule(lr){4-5} \cmidrule(lr){6-7} \cmidrule(lr){8-9}
         
    & & & \hspace{2.4mm}$\text{CHAIR}_s$~$\downarrow$  & \hspace{2.4mm}$\text{CHAIR}_i$~$\downarrow$ & \hspace{2.4mm}Hall.~$\downarrow$ & \hspace{1.4mm}Score~$\uparrow$ & \hspace{2.4mm}Acc.~$\uparrow$ & \hspace{1.4mm}F1~$\uparrow$ \\

    \midrule    
    GPT-4V~\cite{2023GPT4VisionSC}$^\blacktriangle$ & GPT-4V & - & 13.6 & 7.3 & 28.1 & 3.42 & 83.4 & 87.4 \\

    \midrule  
    \rowcolor[rgb]{0.85,0.85,0.85}
    \multicolumn{9}{l}{\textbf{Vanilla LVLMs}} \\
    QWEN-VL~\cite{qwenvl}$^\blacktriangle$  & Qwen-VL-Chat & 10B & 40.4 & 20.7 & 38.5 & 2.76 & 81.9 & 86.4 \\
    
    LLaVA-NeXT~\cite{liu2024llavanext}$^\blacktriangle$ & LLaVA-NeXT & 34B & 12.6 & 6.4 & 34.4 & 3.14 & 81.4 & 85.4 \\

    VCD~\cite{Leng_2024_CVPR_VCD}$^\blacktriangle$ & LLaVA-1.5 & 7B & 48.8 & 24.3 & 54.2 & 2.12 & 71.8 & 74.9 \\
    
    LLaVA-1.5~\cite{li2024llava}     & LLaVA-1.5 & 7B & 52.3 & 25.5 & 52.7 & 2.36 & 73.5 & 77.7 \\

    LLaVA-1.5~\cite{li2024llava}     & LLaVA-1.5 & 13B & 50.7 & 24.8 & 51.4 & 2.39 & 81.8 & 87.3 \\
    
    \midrule
    \rowcolor[rgb]{0.85,0.85,0.85}
    \multicolumn{9}{l}{\textbf{Fine-tuned LVLMs}} \\
    
    LLaVA-RLHF~\cite{llavarlhf}$^\blacktriangle$ & LLaVA-1.5 & 13B & 38.1 & 18.9 & 62.5 & 2.02 & 79.7 & 83.9 \\ 
    
    Silkie~\cite{li2023silkie}$^\blacktriangle$ & Qwen-VL-Chat & 10B & 27.1 & 13.4 & 32.3 & 3.19 & 82.2 & 87.6 \\

    HA-DPO~\cite{ha-dpo}$^\blacktriangle$ & LLaVA-1.5 & 7B & 39.9 & 19.9 & 60.4 & 1.98 & 75.2 & 79.9 \\

    POVID~\cite{povid}$^\blacktriangle$ & LLaVA-1.5 & 7B & 40.4 & 19.1 & 56.2 & 2.08 & \textbf{82.9} & 87.4 \\

    RLHF-V~\cite{yu2024rlhf}$^\blacktriangle$ & Muffin & 13B & 12.2 & 7.5 & 51.0 & 2.45 & 72.6 & 75.0 \\
    
    RLAIF-V~\cite{yu2024rlaif}$^\blacktriangle$ & LLaVA-1.5 & 7B & 8.5 & 4.3 & 29.2 & 3.06 & 76.8 & 84.5 \\

    MDPO~\cite{wang2024mdpo}$^\lozenge$   & LLaVA-1.5 & 7B & 35.7 & 9.8 & 54.0 & 2.39 & 73.4 & 74.7 \\

    DPO (POVID DataSet) & LLaVA-1.5 &  7B  & 48.9 & 22.4 & 56.0 & 2.15 & 75.1 & 78.9 \\

    DPO (RLAIF-V DataSet) & LLaVA-1.5 &  7B  & 19.1  & 9.3 & 36.6 & 2.70 & 76.8 & 81.5 \\
    
    \midrule  
    
    EMPO (POVID DataSet) & LLaVA-1.5 & 7B & 38.1  & 19.3 & 49.1 & 2.58 & \underline{82.7} & 87.1 \\
    
    EMPO (RLAIF-V DataSet) & LLaVA-1.5 & 7B & \textbf{7.16} & \textbf{3.44} & \textbf{25.6} & \textbf{3.21} & 82.4 & \textbf{87.7} \\
          
    \bottomrule  
    \end{tabular}  
    }  
    \setlength{\abovecaptionskip}{4pt}
    \setlength{\belowcaptionskip}{-10pt}
    \centering  
    \caption{Main experimental results. The best and second-best results are highlighted in \textbf{bold} and \underline{underlined}, respectively.
    $^\blacktriangle$ denotes the results are reported by RLAIF-V~\cite{yu2024rlaif}, and $^\lozenge$ denotes the results from the original papers.}  
    
    \label{tab:main_results}  
\end{table*}

\begin{table}[t] 
    \centering   
    \resizebox{\linewidth}{!}{
    \begin{tabular}{l c cc}
    \toprule  
    \multirow{2}{*}{\textbf{Model}} & \textbf{LLaVA-Bench} & \multicolumn{2}{c}{\textbf{MME}} \\
    \cmidrule(lr){2-2} \cmidrule(lr){3-4}
     & overall~$\uparrow$ & Cog.~$\uparrow$ & Per.~$\uparrow$ \\
    \midrule 
    LLaVA-1.5 & \hspace{-3mm}60.6 & \hspace{-3mm}297.5 & \hspace{-2mm}1496.7  \\
    \hspace{1mm} + DPO & \hspace{5mm}66.4\percentred{+9.57\%} & \hspace{5mm}299.3\percentred{+0.61\%} & \hspace{5mm}1356.7\percentgreen{-9.35\%} \\
    \hspace{1mm} + EMPO & \hspace{6mm}69.3\percentred{+14.36\%} & \hspace{5mm}302.8\percentred{+1.78\%} & \hspace{5mm}1389.8\percentgreen{-7.14\%} \\
    \bottomrule  
    \end{tabular}  
    }  
    \setlength{\abovecaptionskip}{4pt}
    \setlength{\belowcaptionskip}{0pt}
    \caption{General capability evaluation results. Red indicates improvements after using EMPO, green indicates performance decline.} 
    \label{tab:common_result}  
    \vspace{-12pt}
\end{table}

\subsection{Experimental settings}
Following~\citeauthor{yu2024rlaif}, we evaluate the models from two aspects: trustworthiness for hallucination and helpfulness for general capability.

For the trustworthiness, we use three benchmarks:
(1) \textbf{CHAIR}~\cite{yu2024rlhf},  a widely used hallucination detection benchmark, which evaluates the multimodal object hallucinations by comparing the generated entities with manually labeled entities in the COCO~\cite{lin2014coco}.
We both report the sentence-level and the entity-level hallucination rate (denoted as $\mathrm{CHAIR}_s$ and $\mathrm{CHAIR}_i$), respectively.
(2) \textbf{MMHal-Bench}~\cite{llavarlhf}, which uses GPT-4 to evaluate model outputs with human responses from two aspects: hallucination rate~(Hall.) and information richness~(Score).
(3) \textbf{AMBER}~\cite{amber}, which evaluates multimodal hallucinations based on 15220 yes-or-no questions. We report accuracy (Acc.) and F1 score on discriminative tasks.

For the helpfulness, we use two benchmarks: (1) \textbf{LLaVA Bench}~\cite{li2024llava} assesses multimodal understanding and reasoning capabilities, with overall accuracy reported across all tasks.
(2) \textbf{MME}~\cite{fu2023mme} evaluates LVLMs on ten perception and four cognition subtasks, with reported scores for both categories (Per. and Cog.).

\paragraph{Baselines.}
We compare our method with state-of-the-art baselines of various types, including general baselines with strong performance and baselines designed to mitigate hallucinations.

First,\textit{Vanilla LVLM baselines.}
We use the open-source LVLMs, i.e., LLaVA-1.5~\cite{li2024llava}, Qwen-VL~\cite{qwenvl}, and LLaVA-Next~\cite{liu2024llavanext} VCD~\cite{Leng_2024_CVPR_VCD} for comparison.
Besides, we also include GPT-4V~\cite{2023GPT4VisionSC} as a closed-source baseline.

Second, \textit{Fine-tuned LVLMs baselines.}
We include seven fine-tuned LVLMs aiming at mitigating hallucinations:
(1) Silkie~\cite{li2023silkie}, which fine-tunes LVLMs using diverse instruction and feedback from GPT-4V.
(2) LLaVA-RLHF~\cite{llavarlhf}, which extends human feedback alignment from text-only models to the multimodal domain.
(3) HA-DPO~\cite{ha-dpo}, which proposes the first multimodal DPO algorithm.
(4) mDPO~\cite{wang2024mdpo}, which optimizes image preferences rather than language preferences to avoid over-optimization issues.
(5) POVID~\cite{zhou2023analyzing}, which fine-tunes VLLMs using model-generated preference data that targets differences between image and text.
(6) RLHF-V~\cite{yu2024rlhf}, which eliminates hallucination of VLLMs using high-quality human feedback to improve precise behavior boundaries.
(7) RLAIF-V~\cite{yu2024rlaif}, which automatically synthesizes preference data and trains the model using iterative DPO.

\paragraph{Training Datasets.}
We use the following preference datasets for training:
(1) \textbf{POVID}~\cite{povid} incorporating 17,000 randomly sampled examples from the LLaVA-Instruct-150K dataset~\cite{liu2023llava}. 
Its hallucinated responses are produced by using GPT-4V~\cite{2023GPT4VisionSC}, which introduces potential errors in areas like object co-occurrence.
(2) \textbf{RLAIF-V}~\cite{yu2024rlaif} is an open-source feedback dataset, including 4,000 instructions from 7 sources, such as MSCOCO~\cite{lin2014mscoco}, Google Landmark v2~\cite{weyand2020google}, and VQA-v2~\cite{goyal2017vqav2}.
Each RLAIF-V instruction pairs with multiple open-source LVLM-generated responses, with more capable LVLMs determining response preferences.

\paragraph{Implementation Details.}
We implement EMPO based on the LLaVA-v1.5-7B~\cite{li2024llava}, which uses CLIP-ViT~\cite{radford2021learning} as the vision module and Vicuna~\cite{zheng2023judging} as the LLM backbone. 
We train the EMPO for 4 epochs using DeepSpeed~\cite{deepspeed}, which is an open-source library by Microsoft for efficient distributed training. 
We set a hyperparameter \( \alpha \) of 0.7 and \( \beta \) of 0.5, an image resolution of 336*336, a learning rate of 5e-7, and a batch size of 8. The training is conducted on 8 A100 GPUs, taking 4 hours on the POVID dataset and approximately 12 hours on the RLAIF-V dataset. 

\begin{figure*}
    \setlength{\abovecaptionskip}{2pt}
    \centering
    \includegraphics[width=1\linewidth]{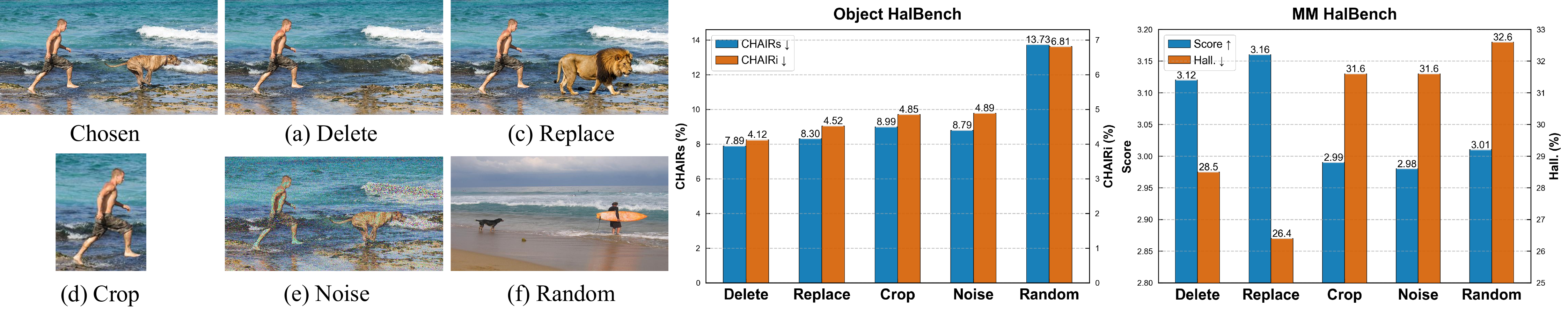}
    \caption{Examples of different rejected image construction strategies and the hallucination rates.}
    \label{fig:rejeted_image}
    \vspace{-5pt}
\end{figure*}

\subsection{Main Results}
\label{main_result}
As shown in Table~\ref{tab:main_results}: 
(1) EMPO can substantially reduce hallucinations in the baseline model LLaVA-1.5-7B. EMPO, trained on either POVID or RLAIF-V, reduces LLaVA-1.5-7B's object hallucination rates on Object HalBench by 24.8\% and 85.9\%, respectively. Additionally, EMPO increases the accuracy (F1) on the AMBER benchmark by 12.5\% (12.9\%) relative percentage points and achieves continuous improvement in terms of overall Hallucination rate on the MMHal dataset. (2) EMPO consistently outperforms DPO across all three benchmarks and two training datasets, with better hallucination reduction effects. This indicates that the proposed EMPO method can effectively improve modal alignment. (3) EMPO achieves state-of-the-art performance in trustworthiness among open-source models, even outperforming commercial models like GPT-4V.

\subsection{Ablation Studies}
Our framework consists of two key components: Multi-modal Preference Alignment and Fine-grained Entity Weighting. 
As shown in Table~\ref{tab: ablations}, to verify the contribution of each component in our framework, we conduct comprehensive ablation studies on the RLAIF-V dataset. 
\paragraph{Ablation of Multi-modal Preference Alignment.}
We first examine the necessity of aligning with human preferences across the \textit{image}, \textit{instruction}, and \textit{response} modalities, respectively.  
Removing any modality preference markedly increases hallucination: +22.9\%, +9.5\%, and +11.9\% for image, instruction, and response, respectively, but still outperforms vanilla DPO. 
The full three-modal alignment performs best, showing that jointly modeling visual and textual signals captures human intent more comprehensively.
\paragraph{Ablation of Fine-grained Entity Weighting.}
To evaluate the effect of explicitly emphasizing key entities, as shown in Table~\ref{tab: ablations}, removing entity weights in Formula~\ref{fomula: weight} increases hallucination by 2.6\%.
Moreover, we study the impact of weighting coefficient $\alpha$ in Formula~\ref{fomula: weight} that balances the entity-centric term and generic preference-alignment loss.
Figure~\ref{fig: alpha_ablation} shows that $\alpha{=}0.7$ yields the lowest hallucination rate on both ObjectHalBench and MMHalBench, indicating that assigning higher weights to entity-related tokens can effectively mitigate hallucination, while the overall semantic coherence of the response cannot be ignored.

\begin{table}[t]
    \setlength{\abovecaptionskip}{4pt}
    \setlength{\belowcaptionskip}{-4pt}
    \centering
    \resizebox{\linewidth}{!}{
    \begin{tabular}{l cc cc}
        \toprule
        \multirow{2}{*}{\textbf{Strategy}} &
        \multicolumn{2}{c}{\textbf{Object-HalBench}} &
        \multicolumn{2}{c}{\textbf{MMHal-Bench}} \\
        \cmidrule(lr){2-3} \cmidrule(lr){4-5}
        & $\text{CHAIR}_{\!s}\,\downarrow$ & $\text{CHAIR}_{\!i}\,\downarrow$ & Hall.\,$\downarrow$ & Res\,$\uparrow$ \\
        \midrule
        Random Instruction & 7.61 & 4.10 & 39.9 & 2.70 \\
        Edit Instruction (EMPO) & \textbf{7.16} & \textbf{3.44} & \textbf{25.6} & \textbf{3.21} \\
        \midrule
        POVID Response & 38.10 & 19.30 & 49.1 & 2.58 \\
        RLAIFV Response (EMPO) & \textbf{7.16} & \textbf{3.44} & \textbf{25.6} & \textbf{3.21} \\
        \bottomrule
    \end{tabular}}
    \caption{Ablation results on different rejected instruction \& response sampling strategies.}
    \label{tab: rejected_text}
\end{table}

\begin{table}[t]
    \setlength{\abovecaptionskip}{4pt}
    \setlength{\belowcaptionskip}{-4pt}
    \centering
    \resizebox{\linewidth}{!}{
    \begin{tabular}{l cc cc}
        \toprule
        \multirow{2}{*}{\textbf{Model}} &
        \multicolumn{2}{c}{\textbf{Object-HalBench}} &
        \multicolumn{2}{c}{\textbf{MMHal-Bench}} \\
        \cmidrule(lr){2-3} \cmidrule(lr){4-5}
        & $\text{CHAIR}_{\!s}\,\downarrow$ & $\text{CHAIR}_{\!i}\,\downarrow$ & Hall.\,$\downarrow$ & Score\,$\uparrow$ \\
        \midrule
        DPO & 19.13 & 9.32 & 36.6 & 2.70 \\
        \midrule
        EMPO & \textbf{7.16} & \textbf{3.44} & \textbf{25.6} & \textbf{3.21} \\
        \hspace{1mm}w/o image       & 10.8 & 5.4 & 34.4 & 2.78 \\
        \hspace{1mm}w/o instruction & 8.5 & 4.1 & 30.9 & 2.96 \\
        \hspace{1mm}w/o response    & 11.2 & 6.3 & 32.1 & 2.87 \\
        \hspace{1mm}w/o weighting   & 7.45 & 3.88 & 28.1 & 3.12 \\
        \bottomrule
    \end{tabular}}
    \caption{Ablation results on different components. We indicates ``w/o image / instruction / response'' denotes removing the corresponding modality preference; ``w/o weighting'' removes the weights on key entities.}
    \label{tab: ablations}
    \vspace{-2pt}
\end{table}

\begin{figure}
    \centering
    \setlength{\abovecaptionskip}{0pt}
    \includegraphics[width=1\linewidth]{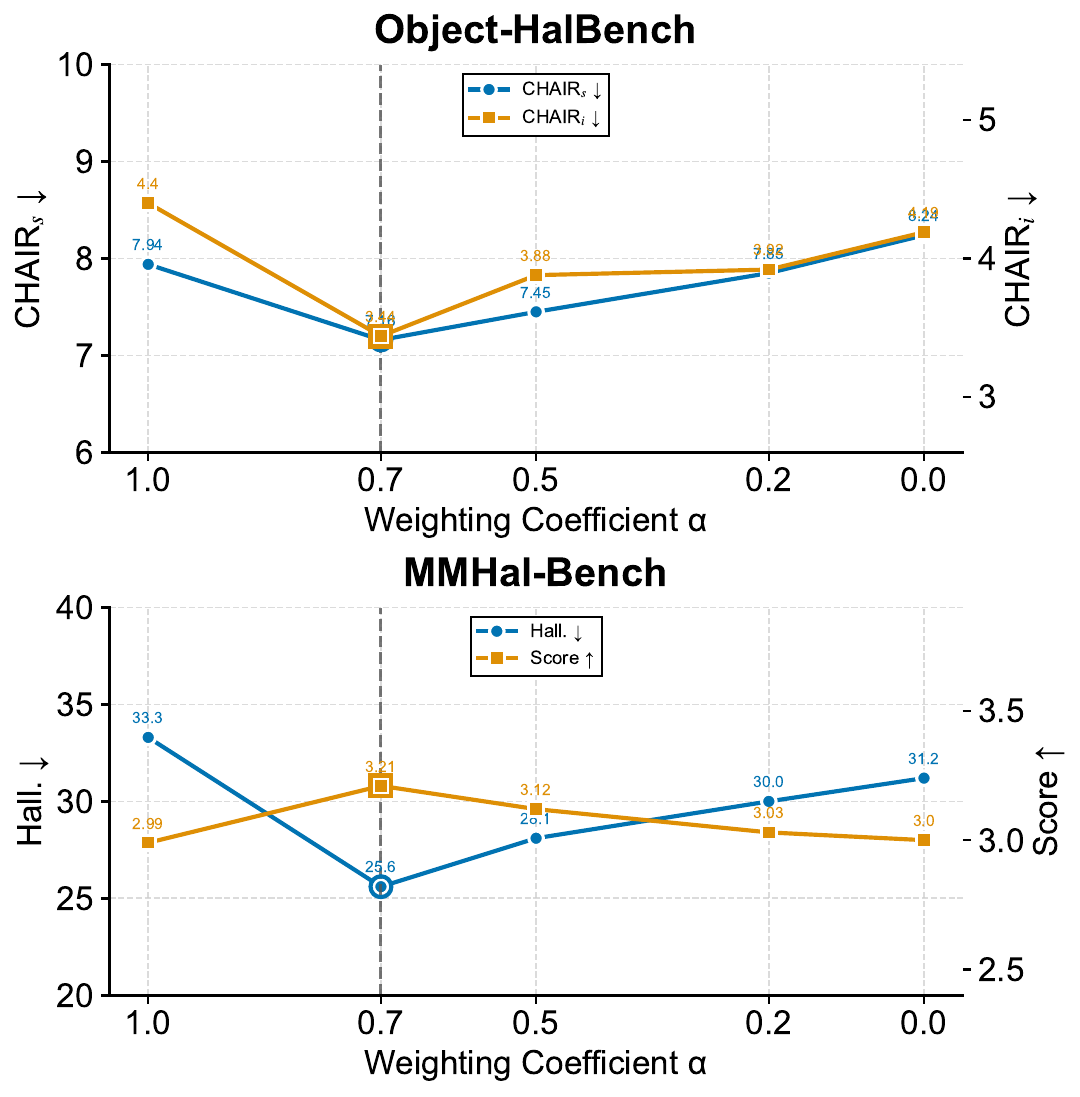}
    \caption{Hyper-parameter ablation on the $\alpha$.}
    \label{fig: alpha_ablation}
    \vspace{-8pt}
\end{figure}
\subsection{Analysis}
\begin{figure*}
    \setlength{\abovecaptionskip}{4pt}
    \centering
    \includegraphics[width=\linewidth]{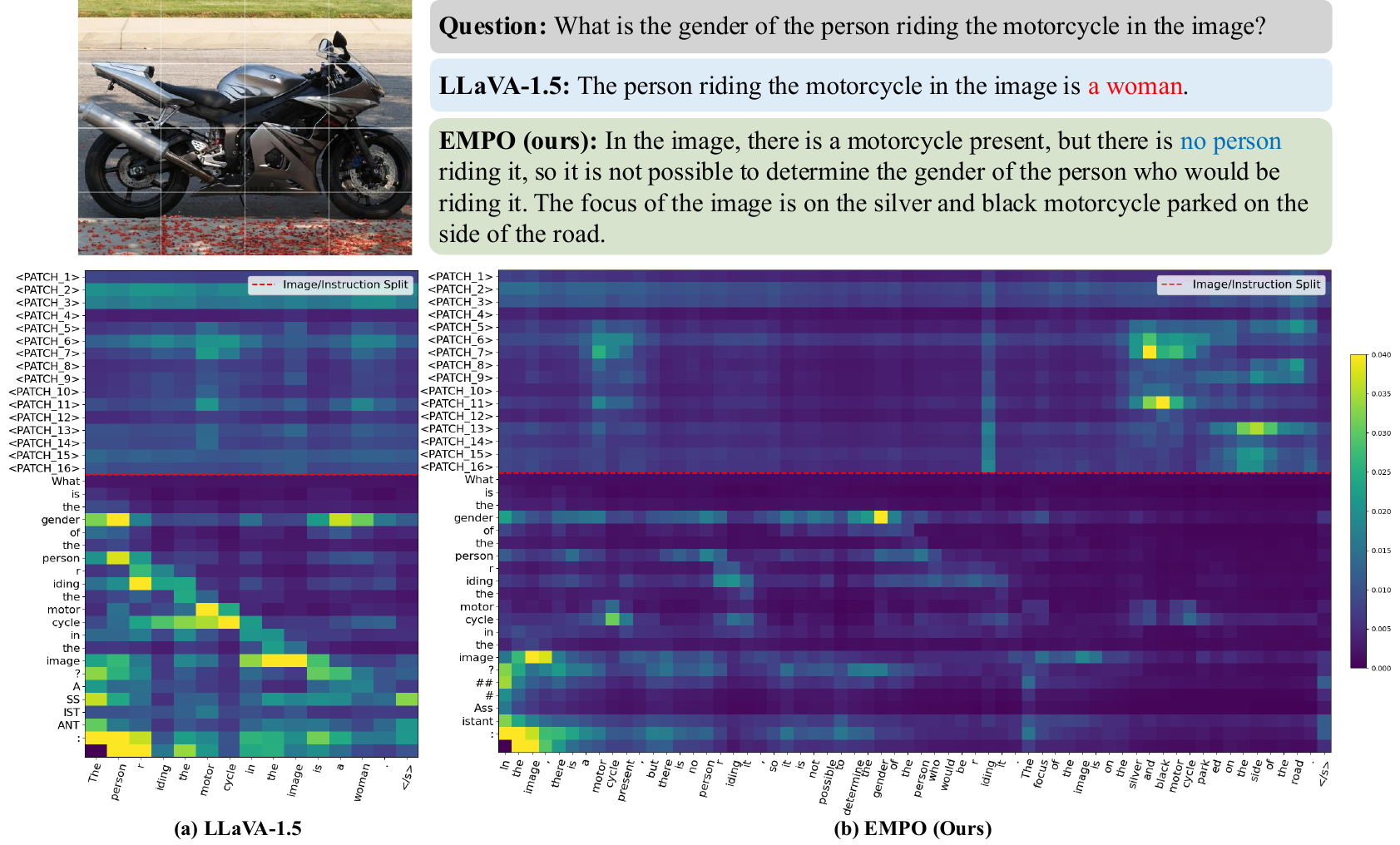}
    \caption{Attention weight heatmaps of outputs to image and instruction tokens from LLaVA-1.5 and EMPO. The hallucinated tokens are highlighted in red. The image is tokenized into 4×4 patches (above the red dash line).}
    \label{fig:case}
    \vspace{-6pt}
\end{figure*}
\paragraph{General Capability Analysis.}
Previous studies show that preference learning may impair models' general understanding capabilities~\cite{xiao2024dposurvey, lan2024survey}. 
We evaluate LVLMs' general capabilities after EMPO training using recognized evaluation datasets: LLaVA-Bench and MME's Perception and Cognition assessments.
Table~\ref{tab:common_result} shows EMPO improves LVLM performance on LLaVA-Bench and MME~(cog.) by 14.36\% and 1.78\% respectively. 
On MME~(Perception), EMPO shows a slight 7.14\% decrease versus the baseline model, though it still outperforms DPO. These results indicate that while EMPO reduces hallucination in LVLMs, it maintains the model's general understanding capabilities.
\paragraph{Impact of Rejection Construction Strategy.}
The construction strategy of preference samples significantly impacts the performance of preference alignment algorithms~\cite{jiang2024survey}. Below, we compare different strategies with our vanilla method in three modalities.

For \textbf{rejected images}, we test five strategies:
(1) Delete: using Stable-Diffusion-2 to remove key entities;
(2) Replace: using Stable-Diffusion-2 to replace key entities with same-type entities;
(3) Crop: randomly cropping the chosen image;
(4) Noise: adding Gaussian noise;
(5) Random: selecting a random training set image.
Figure~\ref{fig:rejeted_image} shows \textit{Delete} and \textit{Replace} outperform others. These strategies maintain structural similarity with chosen images while introducing significant object representation differences, serving as hard negative samples~\cite{kalantidis2020hard} and enabling finer-grained modal feature alignment. We finally adopt a combination of the Delete and Replace as rejected image construction strategies.

For \textbf{rejected instructions}, we test two strategies:
(1) Edit: randomly replacing/deleting entity-related nouns, attributes, and relationships.
(2) Random: selecting a different random instruction from the training set.
As shown in Table~\ref{tab: rejected_text}, Edit achieves a lower hallucination rate than Random, so we select Edit as our primary strategy.

For \textbf{rejected responses}, we experiment with:
(1) POVID~\cite{povid}: using LVLM-generated responses from rejected images.
(2) RLAIF-V~\cite{yu2024rlaif}: constructing preference pairs using an evaluation model.
Table~\ref{tab: rejected_text} shows RLAIF-V significantly outperforms POVID, thus we adopt it for our final EMPO implementation, yet to verify our approach's robustness across different data configurations, we report the results of both instruction datasets compared against DPO in Section~\ref{main_result}.
\paragraph{Modality Alignment Visualization.}\label{sec:visualization}
Figure~\ref{fig:case} presents attention heatmaps illustrating how model output tokens (horizontal axis) attend to input image patches (above the red line) and instruction tokens (below the red line) during inference.
In the baseline LLaVA-1.5 (Figure~\ref{fig:case}(a)), the attention heatmap reveals that while generating the hallucinated entity \emph{"woman"}, the model incorrectly focuses on specific image patches (e.g., <PATCH\_6>, <PATCH\_11>) and the word \emph{"gender"} in the question. 
This misaligned attention pattern exemplifies how the model grounds erroneous assertions in visual features, leading to hallucination.
In contrast, our EMPO-enhanced model (Figure~\ref{fig:case}(b)) demonstrates significantly improved modality alignment. 
When inferring \emph{"no person"} and describing \emph{"silver and black motor"}, the model's attention correctly concentrates on image patches showing only the motorcycle (e.g., <PATCH\_6>, <PATCH\_7>, <PATCH\_11>), effectively verifying the absence of a rider. 
This precise attention allocation shows how EMPO helps LLaVA-1.5 correctly attend to key facts in both the image and user instruction, fostering stronger semantic consistency between visual inputs and generated text to avoid hallucination.

\paragraph{Experiments on more advanced LVLMs.}
\begin{table}[t]
    \setlength{\abovecaptionskip}{4pt}
    \setlength{\belowcaptionskip}{-4pt}
    \centering
    \resizebox{\linewidth}{!}{
    \begin{tabular}{l cc cc}
        \toprule
        \multirow{2}{*}{\textbf{Model}} &
        \multicolumn{2}{c}{\textbf{Object-HalBench}} &
        \multicolumn{2}{c}{\textbf{MMHal-Bench}} \\
        \cmidrule(lr){2-3} \cmidrule(lr){4-5}
        & $\text{CHAIR}_{\!s}\,\downarrow$ & $\text{CHAIR}_{\!i}\,\downarrow$ & Hall.\,$\downarrow$ & Score\,$\uparrow$ \\
        \midrule
        LLaVA-Next(7B)      & 13.4 & 7.0 & 32.9 & 3.08 \\
        \hspace{1mm}+DPO    & 11.9 & 6.8 & 33.6 & 2.99 \\
        \hspace{1mm}+EMPO   & \textbf{5.3}  & \textbf{3.7} & \textbf{29.2} & \textbf{3.18} \\
        \midrule
        Muffin (13B)         & 21.1 & 11.2 & 52.4 & 2.70 \\
        \hspace{1mm}+DPO    & 12.7 & 6.3 & 43.9 & 2.82 \\
        \hspace{1mm}+EMPO   & \textbf{7.0}  & \textbf{4.7} & \textbf{40.8} & \textbf{2.90} \\
        \bottomrule
    \end{tabular}}
    \caption{Experiments on more advanced LVLMs.}
    \label{tab:model_results}
    \vspace{-4pt}
\end{table}

To validate the scalability and effectiveness of EMPO on more advanced LVLMs, we conduct experiments on two state-of-the-art models: LLaVA-Next 7B~\cite{liu2024llavanext} and Muffin 13B~\cite{yu2023muffin}. 
As shown in Table~\ref{tab:model_results}, EMPO substantially reduces hallucination rates across both models, significantly outperforming their vanilla and DPO-trained counterparts. 
Specifically, for LLaVA-Next 7B, EMPO reduces 55.9\% (11.2\%) relative hallucination on Object-HalBench (MMHal-Bench). 
These results underscore EMPO's consistent performance benefits across diverse LVLM architectures.

\section{Conclusion}
\label{sec:conclusion}
This paper addresses the LVLM hallucination problem from two perspectives: modality misalignment and LLM inherent hallucination. 
At the method level, we propose a comprehensive multimodal preference optimization method to help LVLM align entity features with semantic concepts, enhancing its trustworthiness. 
For data side, we introduce a general method for constructing multimodal preference data.
Experiments on multiple benchmarks show our method significantly reduces hallucinations while maintaining LVLM capabilities.
For future work, we will explore common-sense knowledge in multimodal domains and investigate hallucinations in long-term interactive environments like multi-turn dialogue.

\section*{Limitations}
A limitation of this paper is that the investigation into hallucinations was restricted to entity-centric hallucinations. Although entity-centric hallucinations constitute a major component of multimodal hallucinations, non-entity-related hallucinations such as common sense knowledge and long context memory loss are also important aspects for optimizing LVLM effectiveness. Due to space limitations and the complexity of defining common sense and long context, we did not explore these issues in this paper.
We propose the following directions for future research:
(1) Exploring methods to define common-sense knowledge in multimodal domains and its relationship with hallucinations.
(2) Investigating hallucination issues in a long-term interactive environment, such as multi-turn dialogue.

\section*{Ethics Statement}
This study focuses on mitigating hallucination phenomena in LVLMs to enhance their reliability and trustworthiness. We have carefully considered the ethical implications of the research and do not expect any major ethical issues to arise. This study is based on publicly available and widely used data and models; therefore, our findings may inherit the biases and limitations present in these resources.

\section*{Acknowledgments}
We thank all reviewers for their insightful comments and suggestions.
This work is supported by the National Natural Science Foundation of China under Grants 62476188, the Natural Science Foundation of the Jiangsu Higher Education Institutions of China, Key Laboratory of New Generation Artificial Intelligence Technology \& Its
Interdisciplinary Applications (Southeast University), Ministry of Education, China.

\bibliography{custom}

\appendix
\clearpage
\begin{figure*}[h]
    \setlength{\abovecaptionskip}{-12pt}
    \setlength{\belowcaptionskip}{0pt}
    \centering
    \includegraphics[width=1\linewidth]{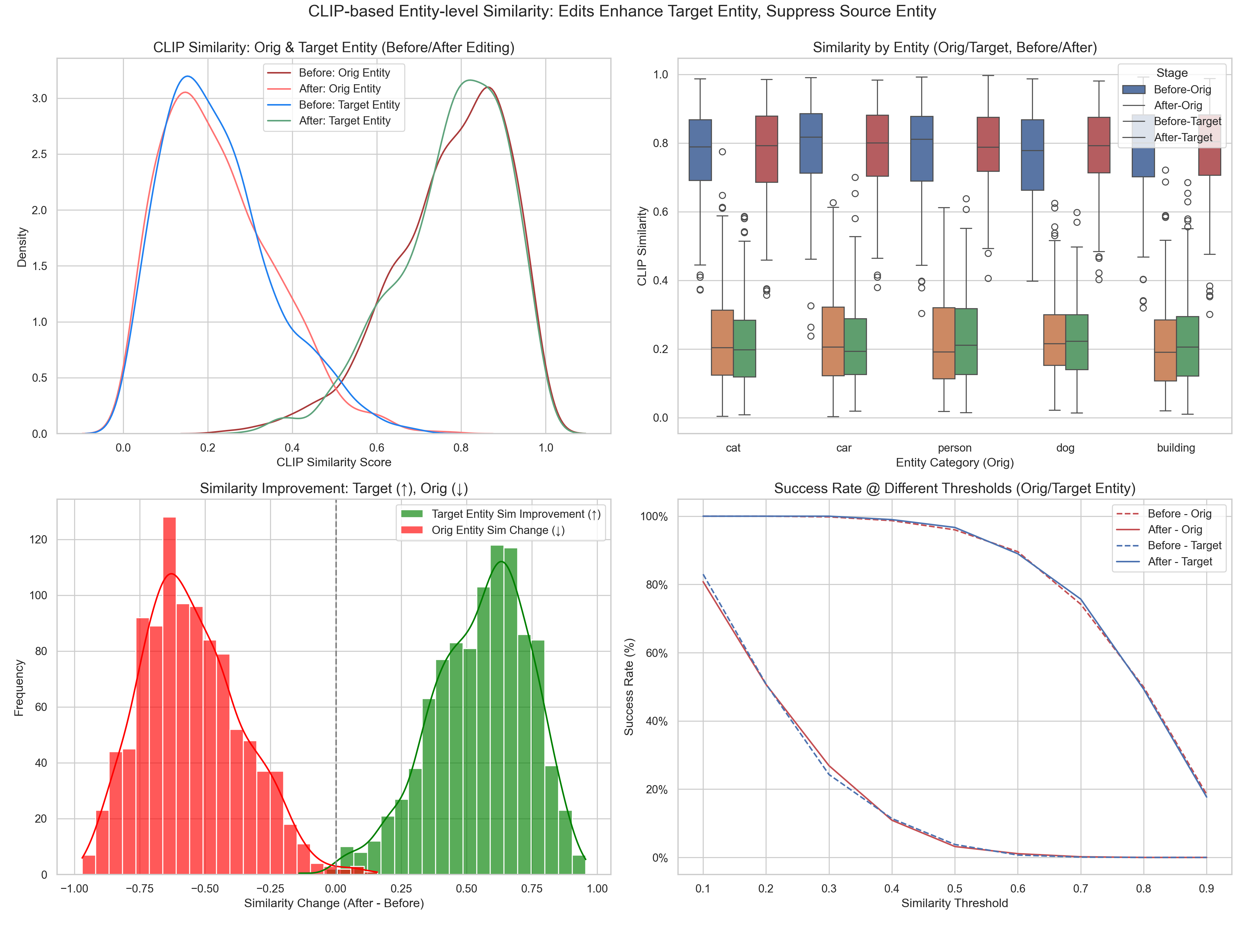}
    \caption{CLIP-based similarity analysis for image editing evaluation: (a) Distribution of similarity scores before and after editing, (b) Box plots showing similarity changes by entity type, (c) Distribution of similarity improvement scores, and (d) Success rates at different similarity thresholds.}
    \label{fig:clip_analysis}
\end{figure*}

\begin{table*}[ht]
    \centering
    \setlength{\abovecaptionskip}{4pt}
    \setlength{\belowcaptionskip}{-4pt}
    \resizebox{0.8\linewidth}{!}{
    \begin{tabular}{l ccc cc cc cc}
        \toprule
        \multirow{2}{*}{\textbf{Model}} & \multicolumn{3}{c}{\textbf{Hallusion-Bench}} & \multicolumn{2}{c}{\textbf{MMBench}} & \multicolumn{2}{c}{\textbf{MMMU}} & \textbf{MMVet} & \textbf{MIA-Bench} \\
        \cmidrule(lr){2-4} \cmidrule(lr){5-6} \cmidrule(lr){7-8} \cmidrule(lr){9-9} \cmidrule(lr){10-10}
        & qA$\uparrow$ & fA$\uparrow$ & aA$\uparrow$ & dev\_en$\uparrow$ & dev\_cn$\uparrow$ & dev$\uparrow$ & test$\uparrow$ & overall$\uparrow$ & overall$\uparrow$ \\
        \midrule
        LLaVA-1.5 & 9.9 & 18.2 & 35.4 & 62.8 & 44.0 & 35.8 & 28.7 & \textbf{32.5} & 67.4 \\
        +EMPO     & \textbf{18.5} & \textbf{21.1} & \textbf{48.2} & \textbf{65.3} & \textbf{47.5} & \textbf{38.3} & \textbf{33.2} & 29.7 & \textbf{71.3} \\
        \bottomrule
    \end{tabular}}
    \caption{Evaluation on more comprehensive benchmarks. We report results on HallusionBench for hallucination assessment, and MMBench, MMMU, MMVet, and MIA-Bench for general capability assessment.}
    \label{tab:comprehensive_results}
    \vspace{-4pt}
\end{table*}

\section{Quality Control and Cost Analysis}
\label{app:data_quality}
\subsection{Quality Control}
We analyze our entity-level editing strategy using CLIP-based semantic similarity, measuring the correspondence of an image region to both source and target entity labels before and after editing. This dual analysis objectively verifies whether the edit enhances the target entity's presence while removing traces of the source.
Using CLIP, we compute similarity scores. As shown in Figure~\ref{fig:clip_analysis}(a), the results reveal a significant shift: post-edit, similarity to the target entity peaks near 0.8, while similarity to the source drops toward zero, reversing the pre-edit pattern. This demonstrates our method’s ability to both install target cues and erase source cues.
This effect is robust across all five entity types (Figure~\ref{fig:clip_analysis}(b)). In every category, the target's median similarity rises post-edit as the source's falls, confirming effective vision-language manipulation. "Person" and "car" categories benefit most, reflecting their strong visual signatures.
Figure~\ref{fig:clip_analysis}(c) quantifies this impact, showing an average similarity increase of 0.6 for the target entity and a corresponding decrease for the source. The minimal overlap between their score distributions indicates reliable substitution. Furthermore, Figure~\ref{fig:clip_analysis}(d) shows that for similarity thresholds above 0.5, over 90\% of edits succeed for the target entity, while the rate for the source approaches zero. This validates that our method produces semantically meaningful transformations.
In summary, our CLIP-based analysis verifies that our pipeline enhances alignment with target entities while suppressing source entity evidence. This fine-grained control supports our claim that our dataset construction achieves clear, contrastive, and controllable entity-centric differences.

\subsection{Cost analysis}
Having established robust quality safeguards, we now consider quantifying the computational requirements and monetary costs of each processing step.
Extracting image entities costs about \$0.002 per entry; generating all 17,332 image entries on 8 A100 GPUs takes roughly 4–6 hours. Instruction-level entity extraction and rewriting add \$0.002 and \$0.003 per entry, respectively. 
In total, the labeling cost is \$0.007 per entry plus 8.3 seconds of compute time on a single A100. Constructing all 17,332 POVID entries cost approximately \$121 for GPT API fees and 40 hours of A100 computing resources (rented through Google Colab, about \$50). By contrast, manual annotation would cost approximately \$0.30 per entry, over 40× higher.

\section{Hallucination Cause Analysis Experiment}
\label{sec:exsupply}
To better understand the severity of hallucinations in LVLMs, we conducted a pilot experiment to evaluate their inference performance. 
Specifically, we assessed the models on 200 preference examples selected from the POVID dataset~\cite{povid}. Through this analysis, we identify two prominent types of errors in LVLM responses, which highlight critical limitations in their reasoning and multimodal understanding capabilities.
\begin{enumerate}
    \item Concept Confusion: We observe that LVLMs struggle to accurately interpret semantic relationships between entities, leading to concept confusion. The models frequently generate identical or highly similar responses to user instructions that were semantically conflicting or conceptually distinct, which suggests that LVLMs may fail to fully grasp the fine-grained differences between related but distinct concepts, resulting in responses that lack precision and contextual appropriateness.
    \item Visual Neglect: When provided with only textual context (i.e., without accompanying visual input), the models tend to generate image-agnostic responses that disregard the potential relevance of visual information. This behavior indicates an over-reliance on textual cues and insufficient attention to visual content, which we attribute to the influence of LLM-induced hallucinations. Such hallucinations appear to bias the models toward text-based reasoning, even in scenarios where visual understanding is critical.
    This is also in line with the previous work PAI~\cite{liu2024paying}
\end{enumerate}


\section{Extensive experiments on more comprehensive benchmarks.}
To further evaluate EMPO's performance, we conduct experiments on five additional benchmarks using the VLMEvalKit~\cite{duan2024vlmevalkit} implementation. 
We use HallusionBench~\cite{guan2024hallusionbench} for hallucination evaluation and four others—MMMU~\cite{yue2024mmmu}, MMBench~\cite{liu2024mmbench}, MMVet~\cite{yu2024mmvet}, and MIA-Bench~\cite{qian2024miabench}—for general capabilities evaluation. 
As shown in Table~\ref{tab:comprehensive_results}, EMPO significantly mitigates hallucination, increasing the HallusionBench \textit{aA} from 35.4\% to 48.2\%. 
It also consistently improves performance on MMBench, MMMU, MMVet, and MIA-Bench. 
These extensive results confirm that EMPO reduces hallucinations while maintaining or enhancing the model's general proficiency.

\section{Prompt Appendix}
\label{sec:prompt}
The section describes the GPT4o-mini prompt for identifying entities, as well as the prompts for rewriting chosen and rejected instructions.

The prompt for rewriting the chosen instruction:
\begin{lstlisting}[basicstyle=\scriptsize\ttfamily, breaklines=true, breakindent=0em, commentstyle=\color{red!50!green!50!blue!50}, frame=shadowbox, rulesepcolor=\color{red!20!green!20!blue!20},numbers=none,literate={`}{\textasciigrave}1]
# prompt for rewriting chosen instruction
prompt = '''Task: Rephrase the following question while maintaining its original meaning:

Original question: {question}

Requirements:
1. If original question was a declarative sentence, then keep rewritten question as a declarative sentence.
2. Ensure the rephrased question is clear, concise, and maintains the original inquiry intent.
3. You may adjust sentence structure or wording, but do not change the essence of the question.
4. If necessary, slightly expand the question to improve clarity, but keep it concise.
5. Use natural, fluent English in the rephrased version.
Please only provide the rephrased question that meets these criteria without any additional explanation.
'''
\end{lstlisting}

The prompt for rewriting rejected instruction:
\begin{lstlisting}[basicstyle=\scriptsize\ttfamily, breaklines=true, breakindent=0em, commentstyle=\color{red!50!green!50!blue!50}, frame=shadowbox, rulesepcolor=\color{red!20!green!20!blue!20},numbers=none,literate={`}{\textasciigrave}1]
# prompt for rewriting rejected instruction
'''You are an expert in creative writing and linguistic transformation. Your task is to rewrite the given question so that its meaning is significantly different from the original, while maintaining the same general structure and format. Follow these guidelines:

1. Analyze the original question's structure, tone, and key elements.
2. Identify a different perspective or context that could radically change the question's meaning.
3. Rewrite the question using the new perspective, ensuring it has a distinctly different meaning.
4. Maintain the original question's format, including any specific phrasing or sentence structure.
5. Ensure the rewritten question is coherent, grammatically correct, and makes sense on its own.

Original question: {question}

Rewritten question:

Provide only the rewritten question without any additional explanation.
'''
\end{lstlisting}

The prompt for identifying entities:
\begin{lstlisting}[basicstyle=\scriptsize\ttfamily, breaklines=true, breakindent=0em, commentstyle=\color{red!50!green!50!blue!50}, frame=shadowbox, rulesepcolor=\color{red!20!green!20!blue!20},numbers=none,literate={`}{\textasciigrave}1]
# prompt for identifying entities
prompt = '''
You are a selective entity replacement engine. You need to perform entity replacement on the original text. 

Core Instructions:  
1. Analyze the input text to identify replaceable entities.  
2. Randomly select approximately 50% of the identified entities for substitution.  
3. Replace the chosen entities with contextually appropriate alternatives.  
4. Maintain grammatical correctness and readability.  
5. Output the modified version and a summary of changes.  

Workflow:  

1. Entity Identification  
   - Named entities (people, places, organizations)  
   - Common nouns  
   - Actions/verbs  
   - Descriptors/adjectives  

2. Replacement Rules:  
   - Maintain the original part of speech.  
   - Preserve sentence structure.  
   - Ensure semantic coherence.  
   - Keep consistent tense and number.  
   - Replace only approximately 50% of the identified entities to retain the original context and flow.  

3. Input Original Text:  
   {original_text}

4. Output Format:  
   - Modified Text:  
     [text with approximately 30% replaced entities]  

   - Changes Summary:  
     - [Original Entity 1] -> [Replacement Entity 1]  
     - [Original Entity 2] -> [Replacement Entity 2]  
     - ...  

Additional Instructions:  

- Entity Selection:
  - After identifying all replaceable entities, calculate 30% of the total number.  
  - Randomly select the calculated number of entities to replace.  
  - Ensure that the selection is random to maintain variability across different texts.  

- Replacement Constraints:
  - Do not replace entities that are crucial for the understanding of the text.  
  - Avoid replacing more than 30% to prevent altering the original meaning significantly.  
  - If the total number of replaceable entities is small, adjust the replacement percentage proportionally to avoid replacing too many.  
'''
\end{lstlisting}
\section{Example Appendix}
\label{sec:example}
This section presents additional examples comparing EMPO and LLaVA-1.5, along with the corresponding heatmaps.

\begin{figure}[h]  
    \centering 
    \begin{minipage}{0.48\textwidth}  
        \centering  
        \includegraphics[width=\linewidth]{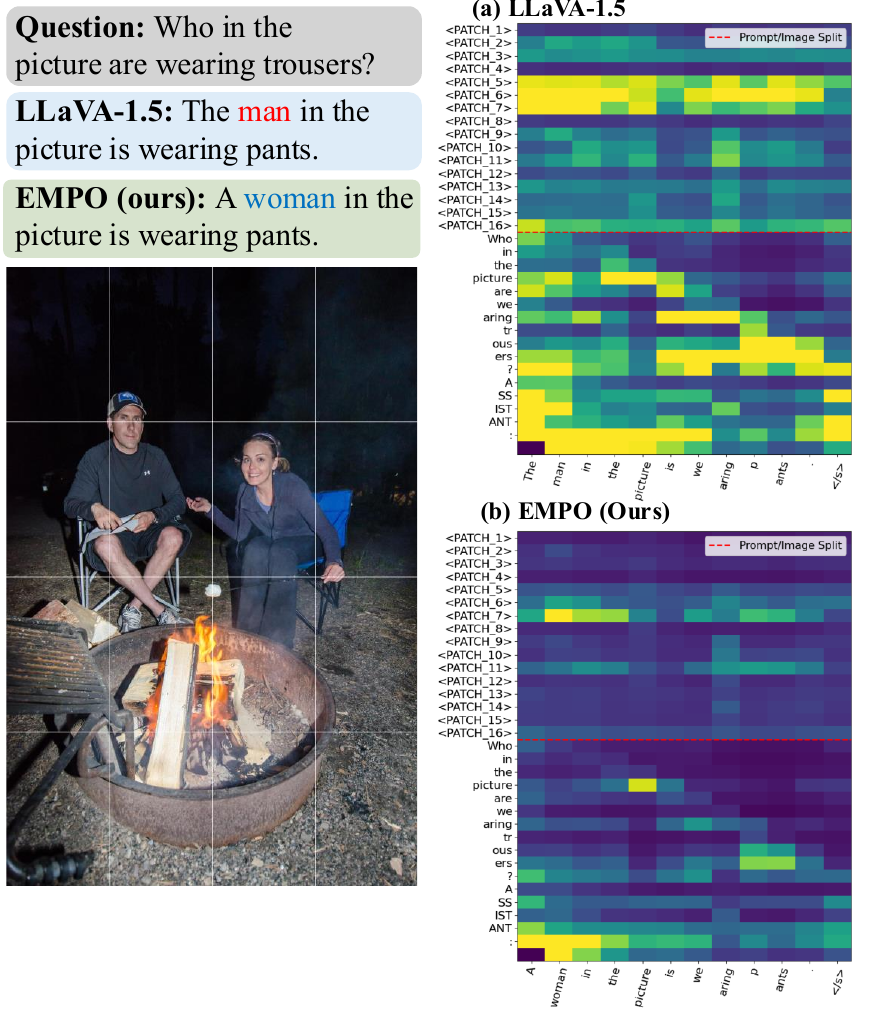}  
        \captionof{figure}{Who in the picture is wearing trousers?}  
    \end{minipage}  
    \hfill 
    \begin{minipage}{0.48\textwidth}  
        \centering  
        \includegraphics[width=\linewidth]{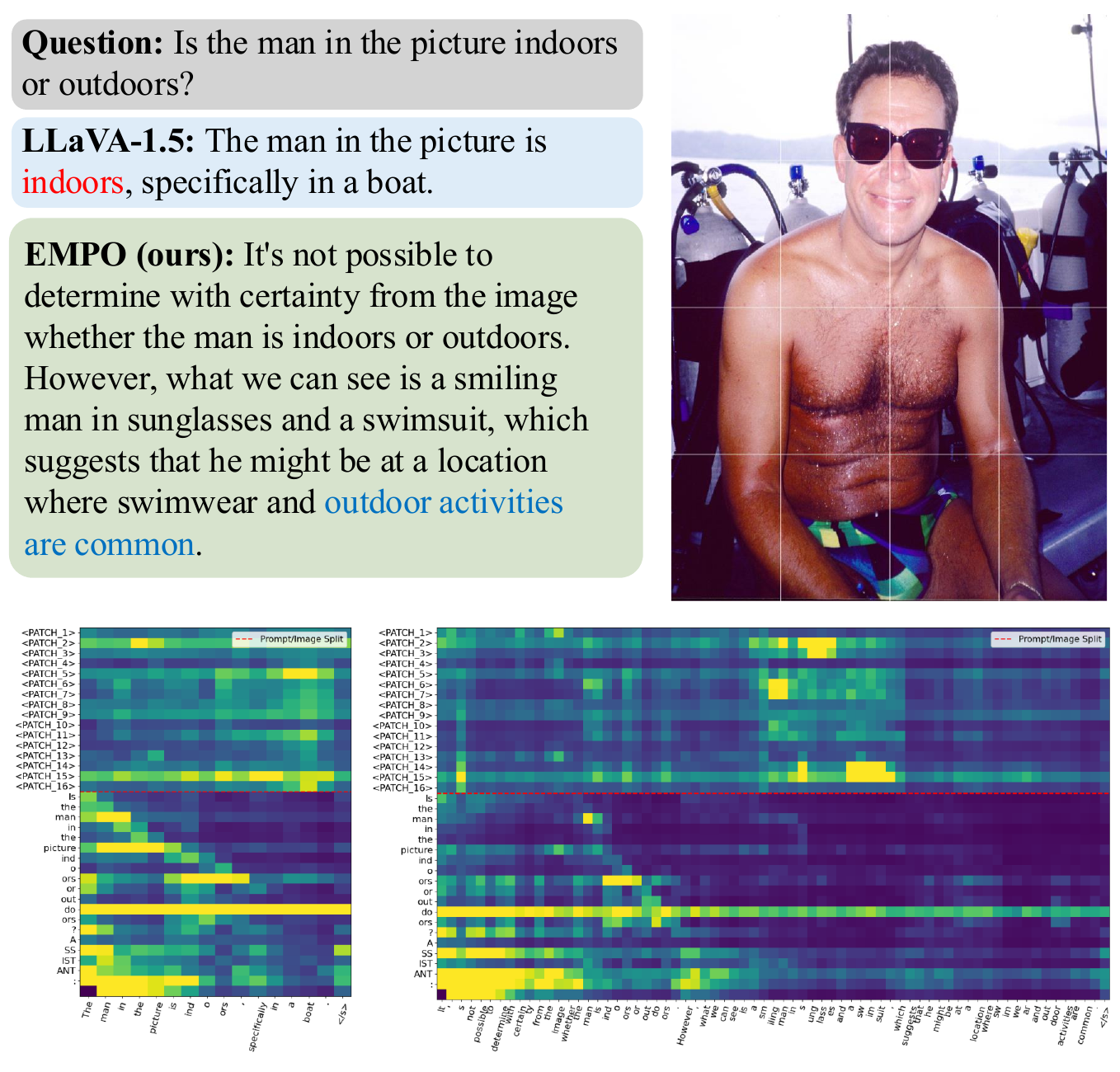}  
        \captionof{figure}{Is the man in the picture indoors or outdoors?}  
    \end{minipage}  
\end{figure} 

\begin{figure*}[h]
    \centering
    \includegraphics[width=\linewidth]{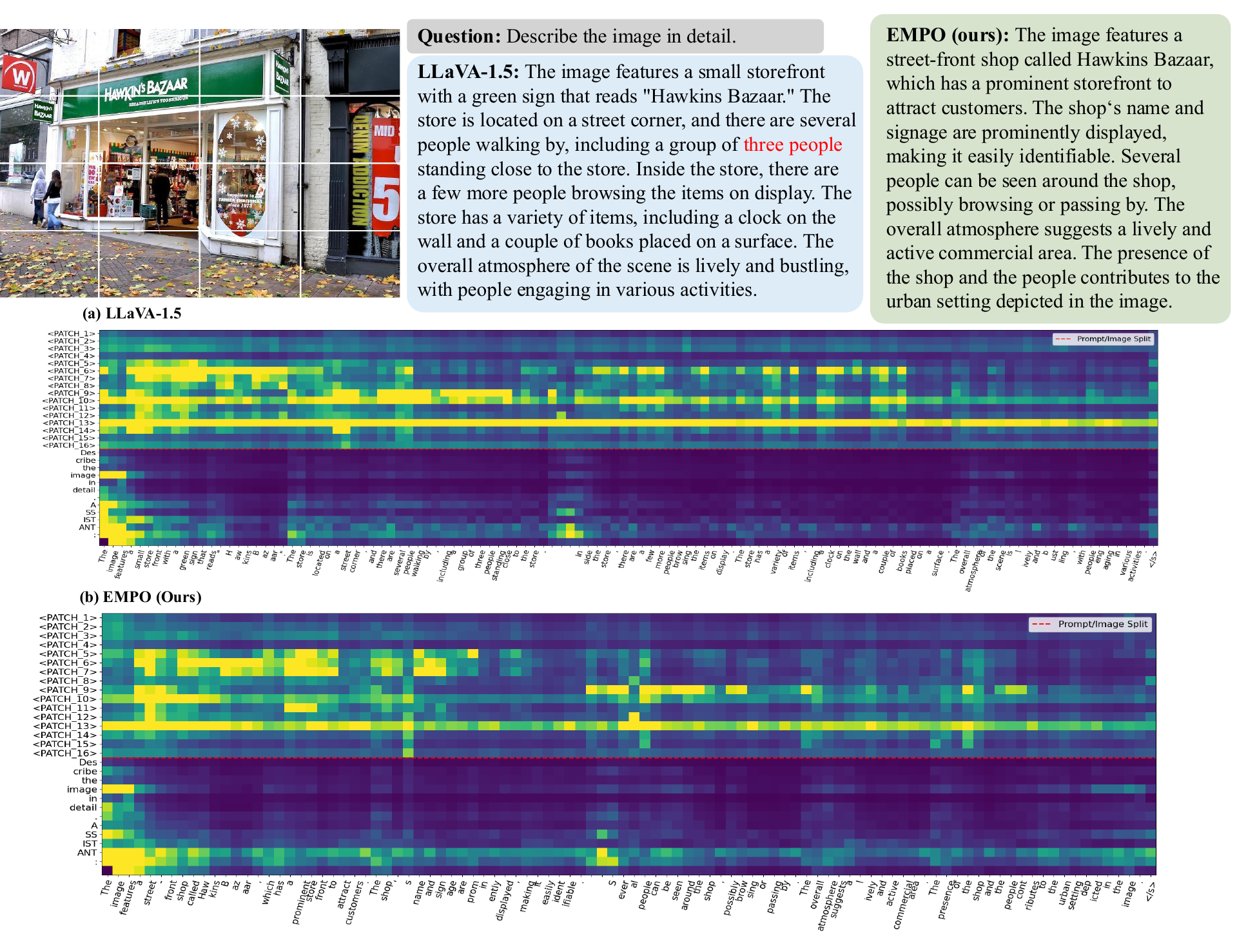}
    \caption{Describe the image in detail.}
\end{figure*}

\begin{figure*}[h]
    \centering
    \includegraphics[width=\linewidth]{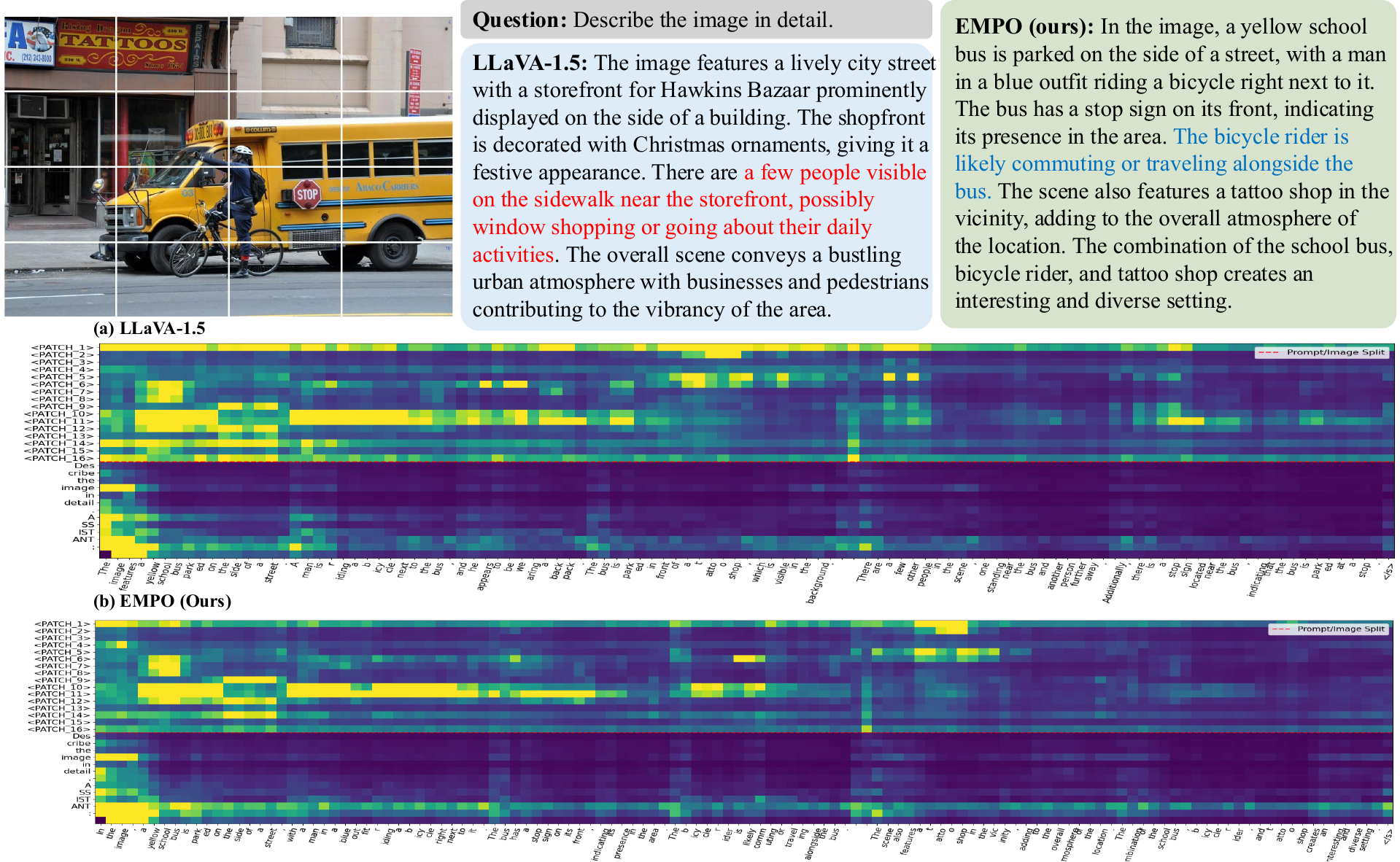}
    \caption{Describe the image in detail.}
\end{figure*}

\end{document}